\PassOptionsToPackage{table}{xcolor}
\documentclass[conference]{IEEEtran}
\IEEEoverridecommandlockouts
\usepackage{pgfplots}
\pgfplotsset{compat=1.17}
\usepackage[numbers]{natbib} 
\usepackage{enumitem}
\usepackage{amsmath,amssymb,amsfonts}
\usepackage{algorithmic}
\usepackage{graphicx}
\usepackage{textcomp}
\usepackage{xcolor}
\usepackage[T1]{fontenc} 

\usepackage{microtype}      
\usepackage{lipsum}
\usepackage{tcolorbox}
\usepackage{tabularx}
\usepackage{booktabs}
\usepackage{booktabs}
\usepackage[table]{xcolor}
\usepackage{multirow}
\usepackage[a4paper, left=0.5in, right=0.5in, top=1in, bottom=1in]{geometry}

\usepackage{graphicx}
\usepackage{tabularx}

\usepackage{forest}
\usetikzlibrary{trees,positioning,shapes,shadows,arrows.meta}
\usepackage{tikz}
\usepackage{hyperref} 
\hypersetup{
    colorlinks=true,
    linkcolor=red,
}

\makeatletter
\newcommand{\linebreakand}{%
  \end{@IEEEauthorhalign}
  \hfill\mbox{}\par
  \mbox{}\hfill\begin{@IEEEauthorhalign}
}
\makeatother

\begin{document}

\title{From Aleatoric to Epistemic: Exploring Uncertainty Quantification Techniques in Artificial Intelligence}

\author{
    \IEEEauthorblockN{
        Tianyang Wang\textsuperscript{a},
        Yunze Wang\textsuperscript{b},
        Jun Zhou\textsuperscript{c},
        Benji Peng\textsuperscript{*, d, l},
        Xinyuan Song\textsuperscript{e},
        Charles Zhang\textsuperscript{d},\\
        Xintian Sun\textsuperscript{f}, 
        Qian Niu\textsuperscript{g}, 
        Junyu Liu\textsuperscript{g}, 
        Silin Chen\textsuperscript{h}, 
        Keyu Chen\textsuperscript{d},
        Ming Li\textsuperscript{d},
        Pohsun Feng\textsuperscript{i}, \\
        Ziqian Bi\textsuperscript{j},
        Ming Liu\textsuperscript{k}, 
        Yichao Zhang\textsuperscript{c},
        Cheng Fei\textsuperscript{m},
        Caitlyn Heqi Yin\textsuperscript{m},
        Lawrence KQ Yan\textsuperscript{n}
    }
    \IEEEauthorblockA{
        \textsuperscript{a}University of Liverpool, UK \\
        \textsuperscript{b}University of Edinburgh, UK \\
        \textsuperscript{c}The University of Texas at Dallas, USA \\
        \textsuperscript{d}Georgia Institute of Technology, USA \\
        \textsuperscript{e}Emory University, USA \\
        \textsuperscript{f}Simon Fraser University, Canada \\
        \textsuperscript{g}Kyoto University, Japan \\
        \textsuperscript{h}Zhejiang University, China \\
        \textsuperscript{i}National Taiwan Normal University, Taiwan \\
        \textsuperscript{j}Indiana University, USA \\
        \textsuperscript{k}Purdue University, USA \\
        \textsuperscript{l}AppCubic, USA \\
        \textsuperscript{m}University of Wisconsin-Madison, USA \\
        \textsuperscript{n}The Hong Kong University of Science and Technology, Hong Kong, China \\
        *Corresponding Email: benji@appcubic.com
    }
}

\maketitle


\begin{abstract}
Uncertainty quantification (UQ) is a critical aspect of artificial intelligence (AI) systems, particularly in high-risk domains such as healthcare, autonomous systems, and financial technology, where decision-making processes must account for uncertainty. This review explores the evolution of uncertainty quantification techniques in AI, distinguishing between aleatoric and epistemic uncertainties, and discusses the mathematical foundations and methods used to quantify these uncertainties. We provide an overview of advanced techniques, including probabilistic methods, ensemble learning, sampling-based approaches, and generative models, while also highlighting hybrid approaches that integrate domain-specific knowledge. Furthermore, we examine the diverse applications of UQ across various fields, emphasizing its impact on decision-making, predictive accuracy, and system robustness. The review also addresses key challenges such as scalability, efficiency, and integration with explainable AI, and outlines future directions for research in this rapidly developing area. Through this comprehensive survey, we aim to provide a deeper understanding of UQ's role in enhancing the reliability, safety, and trustworthiness of AI systems.

\end{abstract}

\section{Introduction}

The widespread application of artificial intelligence (AI) in high-risk fields such as healthcare, autonomous driving, and financial analysis has raised increasing concerns regarding its reliability and safety. AI systems typically operate based on complex models and large amounts of data, and the inherent noise, incompleteness, and limitations of these models lead to unavoidable uncertainty in the system’s outputs. In many application scenarios, failing to effectively quantify and manage this uncertainty may result in severe consequences. For instance, in medical image analysis, neglecting the uncertainty in detecting subtle anomalies in images could lead to misdiagnosis \cite{jiang2017artificial,kurz2022uncertainty}; in autonomous driving, overlooking environmental perception uncertainty may increase the risk of accidents \cite{macfarlane2016addressing}; in the financial sector, failing to account for potential market fluctuations could result in erroneous investment decisions \cite{goodell2020election}. These issues not only affect the accuracy and robustness of AI systems but also deeply influence public trust and acceptance of these technologies. Therefore, how to effectively quantify and control uncertainty has become a critical challenge in current AI research.

Uncertainty in AI can be broadly categorized into two types: aleatoric uncertainty and epistemic uncertainty. Aleatoric uncertainty arises from the intrinsic randomness and noise within the data, such as sensor errors or imprecise measurements. This type of uncertainty is typically irreducible and cannot be eliminated even with more data or improved models \cite{modares2023data}. In contrast, epistemic uncertainty stems from the model’s limitations in understanding the data distribution or environmental changes. It reflects the incompleteness of the model or the lack of sufficient training data to cover all possible scenarios \cite{hullermeier2021aleatoric}. These two types of uncertainty often coexist and interact in real-world applications, requiring approaches that consider their combined effects and apply suitable quantification methods.

In recent years, various uncertainty quantification (UQ) techniques have been proposed, spanning fields such as probabilistic reasoning and deep learning model ensembles. Bayesian inference, as a classical method for handling uncertainty, has been widely applied in deep learning models by incorporating prior distributions to handle uncertainty \cite{abdar2021review}. Sampling-based techniques, such as Monte Carlo methods and dropout, have also been introduced to address uncertainty in deep neural networks \cite{felgentreff2020sampling}. Deep ensemble learning methods, which train multiple models and combine their predictions, further enhance the robustness and accuracy of uncertainty estimates \cite{lakshminarayanan2017simple}. Additionally, generative models like Generative Adversarial Networks (GANs) and Variational Autoencoders (VAEs) have introduced new avenues for modeling uncertainty in data distributions through latent spaces \cite{bohm2019uncertainty}. While these methods have made significant strides in many domains, they still face several challenges, such as high computational complexity, poor real-time performance, and limited adaptability in dynamic environments \cite{bobek2017uncertain}.

Despite the advances made in these techniques, there remain key unresolved issues in the research on uncertainty quantification. For example, how to maintain efficiency and scalability when dealing with large-scale data and complex models \cite{al2019quantifying}, how to improve the applicability of uncertainty measures in multi-modal data \cite{sun2021deep}, and how to integrate uncertainty quantification with interpretability and ethical concerns are critical challenges \cite{mehdiyev2024quantifying}. Moreover, existing approaches often focus on single-model analysis, with a lack of unified frameworks and standardized methods for cross-model or cross-domain applications \cite{han2024approximate}.

This review aims to systematically summarize the progress in uncertainty quantification in AI, focusing on the challenges and solutions in high-risk applications. Specifically, the objectives of this review are as follows:

\begin{itemize}[leftmargin=*] \item Analyze fundamental uncertainty quantification methods, including classical Bayesian reasoning, deep learning methods, ensemble learning, and generative models, discussing their respective advantages, limitations, and appropriate use cases. \item Discuss the application of uncertainty quantification techniques in high-risk fields such as healthcare, autonomous driving, and finance, highlighting their real-world effectiveness and challenges. \item Examine the limitations of current technologies, such as computational complexity, real-time performance, and cross-domain applicability, and propose key technical bottlenecks for future research. \item Explore future research directions, especially in managing uncertainty in large-scale and dynamic environments and integrating uncertainty quantification with explainable AI (XAI), suggesting potential technical pathways. \item Emphasize the role of uncertainty quantification in enhancing the safety, transparency, explainability, and ethical compliance of AI systems, and discuss how these technologies can help build public trust in AI. \end{itemize}

\section{Fundamentals of Uncertainty Quantification}

Uncertainty Quantification (UQ) plays a pivotal role in artificial intelligence (AI) and machine learning (ML), especially when these technologies are applied in high-risk domains such as healthcare, finance, autonomous systems, and engineering \cite{seoni2023application}. In these contexts, the reliability and robustness of AI models depend not only on their accuracy but also on their ability to quantify and handle uncertainty \cite{nemani2023uncertainty}. The ability to assess and reduce uncertainty in model predictions directly influences decision-making processes and enhances the trustworthiness of the system \cite{abdar2022need}. This section provides an in-depth examination of the key principles of UQ, its two primary types—aleatoric and epistemic uncertainty—and the mathematical tools and methods employed to quantify and manage these uncertainties.

\subsection{Types of Uncertainty}

In the field of AI and ML, uncertainty typically arises from two main sources: \textbf{aleatoric uncertainty} and \textbf{epistemic uncertainty}. These two categories of uncertainty represent different underlying causes and have distinct implications for both the modeling process and decision-making \cite{abdar2021review}.

\textbf{Aleatoric Uncertainty} refers to the inherent randomness or noise within a system or dataset. This type of uncertainty stems from unpredictable fluctuations in the data generation process that cannot be eliminated, even with infinite data. Aleatoric uncertainty is typically associated with variability in the system's output due to factors such as measurement errors, inherent variability in natural processes, or stochastic phenomena \cite{hullermeier2021aleatoric}. In regression tasks, for instance, aleatoric uncertainty can be represented as the variance of residual errors. The mathematical representation of aleatoric uncertainty in a simple regression model can be written as:

\[
y = f(x) + \epsilon, \quad \epsilon \sim \mathcal{N}(0, \sigma^2)
\]
Here, \(y\) represents the observed value, \(f(x)\) is the underlying function, and \(\epsilon\) is the noise term, which is assumed to follow a Gaussian distribution with zero mean and variance \(\sigma^2\). This noise term is typically considered irreducible, meaning that it cannot be reduced by collecting more data \cite{oberkampf2004challenge}.

\textbf{Epistemic Uncertainty}, in contrast, arises from a lack of knowledge or insufficient information about the system, the model, or its parameters. Unlike aleatoric uncertainty, epistemic uncertainty is reducible and can be mitigated by obtaining more data, refining model assumptions, or improving the model's representation \cite{stone2022epistemic}. This type of uncertainty is typically associated with the parameters of the model or its structure. For example, in Bayesian inference, epistemic uncertainty is represented by a probability distribution over the model's parameters, reflecting the modeler’s beliefs about the parameters before and after observing data. The formal expression of epistemic uncertainty is through the \textbf{posterior distribution} \(p(\theta|D)\), which represents the updated belief about the parameters \(\theta\) given the observed data \(D\):

\[
p(\theta|D) = \frac{p(D|\theta) p(\theta)}{p(D)}
\]
where \(p(D|\theta)\) is the likelihood function of the data given the model parameters, \(p(\theta)\) is the prior distribution that encodes prior knowledge about the parameters, and \(p(D)\) is the marginal likelihood (also known as the evidence), which normalizes the posterior \cite{hullermeier2021aleatoric}.

\subsection{Mathematical Foundations of UQ}

Uncertainty in AI and ML can be systematically quantified using probability theory and statistics. These mathematical tools provide a framework for modeling uncertainty and making probabilistic predictions. One of the most fundamental concepts in UQ is the \textbf{probability distribution}, which allows us to describe the likelihood of various outcomes for a random variable. The \textbf{probability density function (PDF)} \(p(x)\) provides the likelihood of different values of a continuous random variable \(X\), while the \textbf{cumulative distribution function (CDF)} \(F(x)\) represents the probability that \(X\) takes a value less than or equal to \(x\):

\[
F(x) = \int_{-\infty}^{x} p(x') \, dx'
\]

When analyzing uncertainty, we also make use of \textbf{entropy} as a measure of uncertainty in a probability distribution. The \textbf{Shannon entropy} \(H(X)\) for a discrete random variable \(X\) is defined as:

\[
H(X) = - \sum_{x} p(x) \log p(x)
\]

This quantifies the uncertainty in the distribution: the higher the entropy, the greater the uncertainty. For continuous variables, the differential entropy is used:

\[
H(X) = - \int_{-\infty}^{\infty} p(x) \log p(x) \, dx
\]

Entropy provides a useful metric for comparing uncertainty across different models or datasets, and it is commonly used in information theory and decision theory.

Another key tool in UQ is the \textbf{confidence interval}, which provides a range within which the true value of a parameter or prediction is likely to lie with a given level of confidence (e.g., 95\%). Confidence intervals are widely used in both Bayesian and frequentist statistics to express uncertainty about model predictions \cite{lu2012analysis}.

\subsection{UQ in AI Decision-Making}

In AI systems, uncertainty quantification is essential for making informed decisions under uncertainty. UQ methods allow the system to assess how confident it is in its predictions and guide decision-makers in high-risk environments. For instance, in medical diagnostics, uncertainty quantification can help determine whether the prediction of a disease diagnosis is robust or whether additional data (such as further testing) is necessary \cite{ghoshal2020estimating}.

Decision-making under uncertainty typically involves the use of \textbf{decision theory}, which integrates uncertainty into the optimization of actions. In this framework, decision-makers seek to minimize the expected loss or maximize the expected utility. The expected loss can be expressed mathematically as:

\[
\text{Expected Loss} = \mathbb{E}[L(a, y)] = \sum_{y} p(y|x) L(a, y)
\]
where \(a\) is the action taken (such as recommending a diagnosis), \(y\) is the possible outcome (e.g., true disease status), \(p(y|x)\) is the predicted probability of the outcome, and \(L(a, y)\) is the loss incurred from taking action \(a\) when the true outcome is \(y\). By considering uncertainty in the prediction \(p(y|x)\), the decision-maker can make more robust choices, factoring in the risk associated with each possible outcome.

UQ can also guide \textbf{exploration-exploitation trade-offs} in reinforcement learning (RL) and sequential decision-making problems. In these contexts, models balance between exploring new actions that might reduce uncertainty and exploiting actions that have already been shown to perform well. This trade-off is crucial for improving the model's decision-making process over time.

\subsection{Sources of Uncertainty in AI Systems}

Uncertainty in AI systems can arise from several different sources, each contributing to the overall uncertainty in the system's predictions. The major sources of uncertainty include:

\begin{itemize}
  \item \textbf{Data Uncertainty}: This includes noise in the data, variability in data generation, and missing or incomplete data. Data uncertainty is often modeled as aleatoric uncertainty because it represents inherent randomness in the process that cannot be eliminated through additional data collection \cite{hariri2019uncertainty}.
  \item \textbf{Model Uncertainty}: This arises from limitations in the model itself, including incorrect assumptions about the underlying process, model bias, or the model’s inability to capture all relevant features. Model uncertainty is typically epistemic and can be mitigated through better model design, regularization, and the incorporation of more data \cite{draper1995assessment}.
  \item \textbf{Computational Uncertainty}: AI models, especially deep learning models, involve complex computations that may introduce numerical errors due to finite precision arithmetic or approximations. Stochastic optimization methods, such as stochastic gradient descent (SGD), introduce additional uncertainty due to their random initialization and iterative nature \cite{abdar2021review}.
  \item \textbf{Environmental Uncertainty}: In dynamic systems, such as autonomous vehicles or robotic systems, uncertainty may arise from changes in the environment that the model cannot predict or control. This type of uncertainty can affect both the performance and safety of the system \cite{henne2019managing}.
\end{itemize}

By identifying the sources of uncertainty, AI systems can be designed to account for and mitigate their effects. This is especially important in safety-critical applications, where decision-making under uncertainty is a key factor in ensuring reliability and minimizing risk.

\section{Advances in Uncertainty Quantification Techniques}

Uncertainty Quantification (UQ) plays a critical role in improving the robustness, interpretability, and reliability of machine learning (ML) systems, particularly in high-stakes applications such as healthcare, finance, and autonomous systems. Advances in UQ methods have significantly broadened their applicability, enabling nuanced characterization of uncertainties across diverse tasks and domains. This section provides an in-depth exploration of contemporary UQ techniques, classified into six primary categories: probabilistic methods, ensemble learning methods, sampling-based approaches, generative models, deterministic methods, and emerging hybrid techniques.

\subsection{Probabilistic Methods}

Probabilistic methods form the foundation of UQ by representing uncertainties using probability distributions, which provide interpretable metrics such as mean, variance, and confidence intervals. Bayesian approaches, particularly Bayesian Neural Networks (BNNs), are central to this paradigm. BNNs incorporate priors over model parameters \( p(\theta) \) and update these priors using observed data \(\mathcal{D}\) to compute the posterior distribution \( p(\theta|\mathcal{D}) \). The predictive distribution, which reflects both epistemic and aleatoric uncertainties, is given by:

\begin{equation}
p(y|x, \mathcal{D}) = \int p(y|x, \theta) p(\theta|\mathcal{D}) d\theta.
\end{equation}

Exact inference for BNNs is computationally prohibitive, necessitating the use of approximation techniques like Variational Inference (VI) and Monte Carlo (MC) Dropout. VI optimizes a simpler variational distribution \( q(\theta) \) to approximate \( p(\theta|\mathcal{D}) \) by minimizing the Kullback-Leibler (KL) divergence:

\begin{equation}
\text{KL}(q(\theta) \| p(\theta|\mathcal{D})).
\end{equation}

Applications of probabilistic methods include predictive modeling in clinical settings \cite{johnson2016mimic}, financial forecasting \cite{tsay2005analysis}, and autonomous decision-making \cite{thrun2002probabilistic}. Despite their utility, these methods face challenges such as scalability to large datasets \cite{dean2008mapreduce}, sensitivity to prior selection \cite{gelman1995bayesian}, and computational overhead \cite{kingma2013auto}.

\subsection{Ensemble Learning Methods}

Ensemble methods leverage the diversity among multiple models to estimate uncertainty. In \textit{deep ensembles}, several neural networks are independently trained with different initial conditions or subsets of training data, and their predictions are aggregated to compute the mean and variance \cite{lakshminarayanan2017simple}:

\begin{equation}
p(y|x) = \frac{1}{M} \sum_{i=1}^M p(y|x, \theta_i),
\end{equation}
where \( M \) is the number of models, and \( \theta_i \) represents the parameters of the \( i \)-th model. This approach captures \textit{aleatoric uncertainty}, arising from data noise, and \textit{epistemic uncertainty}, due to model limitations or insufficient data.

Deep ensembles are particularly effective for tasks involving safety-critical decisions, such as medical image diagnosis or autonomous navigation, offering robustness against adversarial perturbations. However, the computational cost and memory requirements of training and storing multiple models remain key drawbacks. Efforts to address these limitations include distillation-based ensemble approximations \cite{hinton2015distilling} and shared-weight architectures \cite{liang2021pruning}.

\subsection{Sampling-Based Methods}

Sampling-based methods are among the most flexible UQ approaches, capable of approximating complex posterior distributions through stochastic sampling. \textit{Monte Carlo (MC) Sampling} generates predictions by repeatedly sampling from the posterior distribution, allowing the computation of metrics such as mean and variance. For example, MC Dropout uses multiple stochastic forward passes with dropout enabled, providing an estimate of uncertainty through the variance of predictions:

\begin{equation}
\text{Var}(y|x) \approx \frac{1}{T} \sum_{t=1}^T (\hat{y}_t - \bar{y})^2,
\end{equation}
where \( T \) is the number of samples, \( \hat{y}_t \) is the \( t \)-th prediction, and \( \bar{y} \) is the mean prediction.

Advanced techniques, such as Hamiltonian Monte Carlo (HMC) \cite{neal2012mcmc} and Sequential Monte Carlo (SMC) \cite{doucet2001sequential}, improve sampling efficiency. HMC incorporates gradient information to explore the posterior more effectively, while SMC updates posterior samples sequentially, making it suitable for time-evolving systems. These methods are particularly valuable in Bayesian optimization \cite{snoek2012practical}, model calibration, and uncertainty-aware reinforcement learning \cite{osband2016deep}.

\subsection{Generative Models}

Generative models have emerged as powerful tools for UQ by learning data distributions and providing uncertainty estimates through latent representations. \textit{Variational Autoencoders (VAEs)}, for instance, learn a probabilistic mapping between observed data and latent variables \( z \), optimizing the evidence lower bound (ELBO) \cite{kingma2013auto}:

\begin{equation}
\mathcal{L}_{\text{VAE}} = \mathbb{E}_{q_\phi(z|x)}[\log p_\theta(x|z)] - \text{KL}(q_\phi(z|x) \| p(z)).
\end{equation}

\textit{Generative Adversarial Networks (GANs)}, extended to Bayesian GANs, incorporate uncertainty in their generative processes, making them suitable for data synthesis and outlier detection \cite{saatci2017bayesian}. \textit{Normalizing Flows}, with their exact likelihood computation, transform simple base distributions into complex ones, providing fine-grained uncertainty estimates \cite{rezende2015variational}.

Generative models are widely applied in medical imaging \cite{chen2018deep}, physics-informed modeling \cite{yang2020physics}, and anomaly detection \cite{schlegl2017unsupervised}. Despite their versatility, challenges such as mode collapse in GANs \cite{che2016mode} and the sensitivity of VAEs to hyperparameter settings \cite{burgess2018understanding} require careful design and tuning.

\subsection{Deterministic Methods}

Deterministic approaches provide alternative strategies for UQ, emphasizing computational efficiency and interpretability. \textit{Evidential Deep Learning (EDL)} models the uncertainty of classification tasks using Dirichlet distributions, parameterized by evidence variables derived from model outputs \cite{sensoy2018evidential}:

\begin{equation}
p(y|x) = \int \text{Dir}(\alpha) p(\alpha|x) d\alpha.
\end{equation}

Interval-based methods, such as Quantile Regression, predict confidence intervals directly, providing bounds on outputs without requiring stochastic sampling \cite{koenker1978regression}. These methods are particularly attractive for real-time applications or resource-constrained environments. Although deterministic methods are efficient and straightforward to implement, they may lack the flexibility to capture complex uncertainty structures, especially in multimodal or high-dimensional problems.

\subsection{Others}

Emerging techniques in UQ explore hybrid models and domain-specific approaches. \textit{Hybrid Methods} integrate multiple UQ strategies, such as combining Bayesian inference with ensemble models or embedding deterministic methods within probabilistic frameworks. \textit{Physics-Informed Neural Networks (PINNs)} impose domain-specific physical constraints, ensuring consistency with known laws and reducing uncertainty in scientific applications \cite{raissi2019physics}.

Information-theoretic measures, such as mutual information \( \mathbb{I}(y; \theta|x) \), are increasingly used to quantify epistemic uncertainty in active learning and decision-making tasks:

\begin{equation}
\mathbb{I}(y; \theta|x) = H(p(y|x)) - \mathbb{E}_{p(\theta|\mathcal{D})}[H(p(y|x, \theta))].
\end{equation}

These emerging approaches have demonstrated promise in areas such as robotics, climate science, and material discovery, where uncertainty quantification must integrate domain knowledge and computational constraints \cite{settles2009active}.

\section{Evaluation Metrics for Uncertainty Quantification}

The evaluation of Uncertainty Quantification (UQ) methods is crucial to validate their effectiveness in capturing, representing, and leveraging uncertainty in predictive tasks. Metrics for UQ address multiple dimensions, including calibration, sharpness, reliability, and practical utility across different tasks. This section provides a detailed discussion of these evaluation metrics, emphasizing mathematical rigor and practical considerations.

\subsection{Calibration Metrics}
Calibration reflects how well predicted uncertainties match observed outcomes. A calibrated model ensures that its predicted probabilities or confidence intervals align with actual event frequencies, enhancing reliability \cite{guo2017calibration}.

\paragraph{Expected Calibration Error (ECE)}
ECE is a widely used metric that aggregates the calibration error across multiple confidence bins. It quantifies the average deviation between predicted confidence and actual accuracy \cite{naeini2015obtaining}:
\begin{equation}
\text{ECE} = \sum_{m=1}^M \frac{|B_m|}{n} \left| \text{acc}(B_m) - \text{conf}(B_m) \right|,
\end{equation}
where $M$ is the number of bins, $B_m$ is the set of predictions in bin $m$, $|B_m|$ is the size of the bin, $n$ is the total number of samples, $\text{acc}(B_m)$ is the accuracy within the bin, and $\text{conf}(B_m)$ is the mean predicted confidence.

\paragraph{Maximum Calibration Error (MCE)}
MCE identifies the maximum calibration error across bins \cite{wentzell1997maximum}:
\begin{equation}
\text{MCE} = \max_{m \in \{1, \ldots, M\}} \left| \text{acc}(B_m) - \text{conf}(B_m) \right|.
\end{equation}

\paragraph{Reliability Diagrams}
A reliability diagram is a graphical tool for assessing calibration. It plots predicted confidence ($x$-axis) against observed accuracy ($y$-axis). A perfectly calibrated model corresponds to a diagonal line, and deviations from this line indicate calibration errors.

\paragraph{Brier Score}
The Brier score measures the accuracy of probabilistic predictions in classification tasks by computing the mean squared error between predicted probabilities ($p_i$) and actual outcomes ($y_i$):
\begin{equation}
\text{Brier Score} = \frac{1}{n} \sum_{i=1}^n (p_i - y_i)^2.
\end{equation}

\subsection{Sharpness Metrics}
Sharpness assesses the concentration of the predictive distribution, independent of its calibration \cite{mitchell2011evaluating}. It is a measure of how confident the predictions are, with sharper predictions being desirable if they remain accurate and calibrated.

\paragraph{Prediction Interval Width (PIW)}
In regression tasks, PIW evaluates the sharpness of confidence intervals:
\begin{equation}
\text{PIW} = \frac{1}{n} \sum_{i=1}^n (U_i - L_i),
\end{equation}
where $U_i$ and $L_i$ represent the upper and lower bounds of the predicted confidence interval for the $i$-th sample.

\paragraph{Entropy}
For classification tasks, predictive entropy quantifies the uncertainty inherent in the predictions:
\begin{equation}
\mathbb{H}(p(y|x)) = -\sum_{k=1}^K p(y=k|x) \log p(y=k|x),
\end{equation}
where $K$ is the number of classes, and $p(y=k|x)$ is the predicted probability for class $k$.

\subsection{Scoring Rules}
Scoring rules provide a unified framework to evaluate predictive distributions by combining calibration and sharpness into a single metric.

\paragraph{Logarithmic Score (Log Score)}
The log score measures the likelihood of observed outcomes under the predicted distribution:
\begin{equation}
\text{Log Score} = -\frac{1}{n} \sum_{i=1}^n \log p(y_i|x_i),
\end{equation}
where $p(y_i|x_i)$ is the predicted probability (or density) of the true outcome $y_i$.

\paragraph{Continuous Ranked Probability Score (CRPS)}
CRPS evaluates probabilistic predictions by comparing the predicted cumulative distribution function (CDF) to the true outcome:
\begin{equation}
\text{CRPS} = \frac{1}{n} \sum_{i=1}^n \int_{-\infty}^\infty \left[ F(x) - \mathbb{I}(x \geq y_i) \right]^2 dx,
\end{equation}
where $F(x)$ is the predicted CDF and $\mathbb{I}$ is the indicator function.

\subsection{Task-Specific Metrics}
Task-specific metrics are tailored to the requirements of specific applications, providing domain-relevant insights into UQ performance.

\paragraph{Coverage Probability}
For regression tasks, the coverage probability assesses the fraction of true outcomes that fall within the predicted confidence intervals:
\begin{equation}
\text{Coverage} = \frac{1}{n} \sum_{i=1}^n \mathbb{I}(y_i \in [L_i, U_i]),
\end{equation}
where $[L_i, U_i]$ is the confidence interval for the $i$-th prediction.

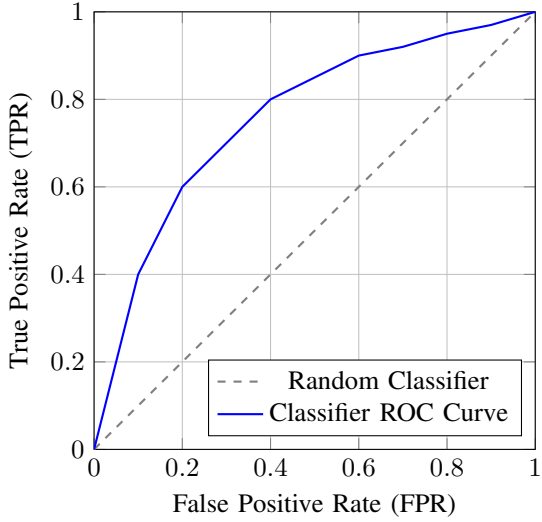
\begin{figure}[ht]
    \centering
    \begin{tikzpicture}
        \begin{axis}[
            width=0.4\textwidth,
            height=0.4\textwidth,
            xlabel={False Positive Rate (FPR)},
            ylabel={True Positive Rate (TPR)},
            xmin=0, xmax=1,
            ymin=0, ymax=1,
            grid=both,
            grid style={line width=.1pt, draw=gray!10},
            major grid style={line width=.2pt,draw=gray!50},
            legend pos= south east
        ]
        \addplot [thick, dashed, gray] coordinates {(0,0) (1,1)};
        \addlegendentry{Random Classifier}
        
        \addplot [thick, blue] coordinates {
            (0,0) (0.1,0.4) (0.2,0.6) (0.3,0.7) (0.4,0.8) (0.5,0.85) (0.6,0.9) (0.7,0.92) (0.8,0.95) (0.9,0.97) (1,1)
        };
        \addlegendentry{Classifier ROC Curve}
        \end{axis}
    \end{tikzpicture}
    \caption{Receiver Operating Characteristic (ROC) Curve illustrating the trade-off between TPR and FPR. The Area Under the ROC Curve (AUROC) quantifies the overall ability of the classifier to discriminate between classes.}
    \label{fig:auroc}
\end{figure}

\paragraph{Area Under the Receiver Operating Characteristic Curve (AUROC)}  
For out-of-distribution (OOD) detection, AUROC evaluates the ability of uncertainty scores to distinguish between in-distribution and OOD samples. As depicted in Fig.~\ref{fig:auroc}, the ROC curve illustrates the trade-off between the True Positive Rate (TPR) and the False Positive Rate (FPR) at various threshold settings. The Area Under the ROC Curve (AUROC) quantifies the overall ability of the classifier to discriminate between classes, with a higher AUROC indicating better performance \cite{fawcett2006introduction}.

\subsection{Comparative and Visualization Techniques}
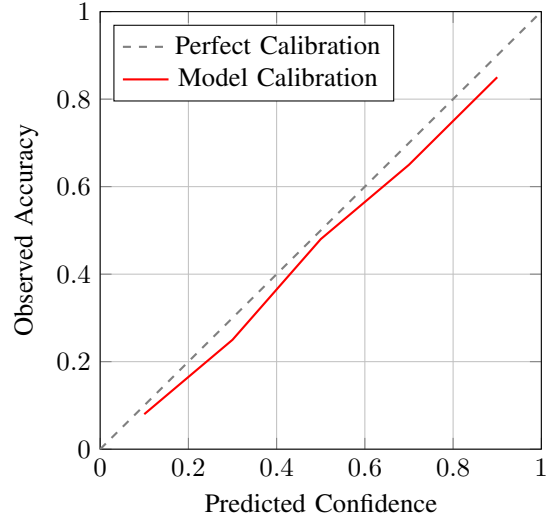
\begin{figure}[ht]
    \centering
    \begin{tikzpicture}
        \begin{axis}[
            width=0.4\textwidth,
            height=0.4\textwidth,
            xlabel={Predicted Confidence},
            ylabel={Observed Accuracy},
            xmin=0, xmax=1,
            ymin=0, ymax=1,
            grid=both,
            grid style={line width=.1pt, draw=gray!10},
            major grid style={line width=.2pt,draw=gray!50},
            legend pos= north west
        ]
        \addplot [thick, dashed, gray] coordinates {(0,0) (1,1)};
        \addlegendentry{Perfect Calibration}
        
        \addplot [thick, red] coordinates {
            (0.1,0.08) (0.3,0.25) (0.5,0.48) (0.7,0.65) (0.9,0.85)
        };
        \addlegendentry{Model Calibration}
        \end{axis}
    \end{tikzpicture}
    \caption{Calibration Plot showing the relationship between predicted confidence and observed accuracy. The closer the calibration curve is to the diagonal line, the better the model is calibrated.}
    \label{fig:calibration}
\end{figure}

\paragraph{Uncertainty Calibration Plots}  
Calibration plots visualize the relationship between predicted confidence and observed outcomes, highlighting systematic biases in uncertainty estimates. As depicted in Fig.~\ref{fig:calibration}, a well-calibrated model aligns closely with the diagonal line, indicating that predicted confidences match observed accuracies \cite{guo2017calibration}.

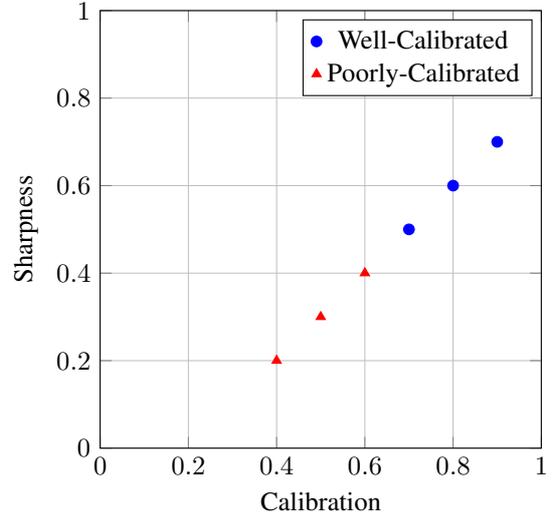
\begin{figure}[ht]
    \centering
    \begin{tikzpicture}
        \begin{axis}[
            width=0.4\textwidth,
            height=0.4\textwidth,
            xlabel={Calibration},
            ylabel={Sharpness},
            xmin=0, xmax=1,
            ymin=0, ymax=1,
            grid=both,
            grid style={line width=.1pt, draw=gray!10},
            major grid style={line width=.2pt,draw=gray!50},
            scatter/classes={
                a={mark=*,blue},
                b={mark=triangle*,red}
            }
        ]
        \addplot[scatter,only marks,scatter src=explicit symbolic]
            coordinates {
                (0.9,0.7) [a]
                (0.8,0.6) [a]
                (0.7,0.5) [a]
                (0.6,0.4) [b]
                (0.5,0.3) [b]
                (0.4,0.2) [b]
            };
        \legend{Well-Calibrated, Poorly-Calibrated}
        \end{axis}
    \end{tikzpicture}
    \caption{Trade-Off between Calibration and Sharpness. Models aim to achieve high sharpness while maintaining good calibration. Points represent different models or configurations.}
    \label{fig:tradeoff}
\end{figure}

\paragraph{Sharpness vs. Calibration Trade-Offs}  
Balancing sharpness and calibration is crucial. As depicted in Fig.~\ref{fig:tradeoff}, models with overly sharp predictions may sacrifice calibration, while overly calibrated models may produce overly wide intervals \cite{naeini2015obtaining}.

\begin{figure}[ht]
    \centering
    \begin{tikzpicture}
        \begin{axis}[
            width=0.4\textwidth,
            height=0.4\textwidth,
            xlabel={Input Feature},
            ylabel={Predicted Output},
            xmin=0, xmax=10,
            ymin=-1.5, ymax=1.5,
            grid=both,
            grid style={line width=.1pt, draw=gray!10},
            major grid style={line width=.2pt,draw=gray!50},
            legend pos= north west
        ]
        \addplot [thick, black, domain=0:10, samples=100] {sin(deg(x))};
        \addlegendentry{True Function}
        
        \addplot [thick, blue, domain=0:10, samples=100] {sin(deg(x)) + 0.1*rand};
        \addlegendentry{Predicted Mean}
        
        \addplot [thick, red, dashed, domain=0:10, samples=100] {sin(deg(x)) + 0.3};
        \addplot [thick, red, dashed, domain=0:10, samples=100] {sin(deg(x)) - 0.3};
        \addlegendentry{Confidence Interval}
        \end{axis}
    \end{tikzpicture}
    \caption{Regression Plot with Predicted Confidence Intervals. The blue line represents the predicted mean, while the red dashed lines denote the confidence intervals. The true function is shown in black.}
    \label{fig:confidence_intervals}
\end{figure}
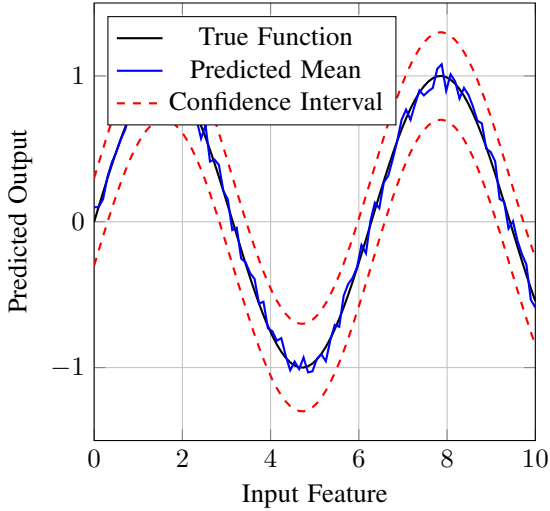

\paragraph{Visualizing Confidence Intervals}  
For regression tasks, plotting predicted intervals against true values provides insights into both sharpness and coverage. As illustrated in Fig.~\ref{fig:confidence_intervals}, the predicted mean and confidence intervals can be compared against the true function to assess the model's uncertainty estimates \cite{koenker1978regression}.

\section{Applications of Uncertainty Quantification in AI}

In safety-critical and high-stakes domains, UQ provides essential insights into the confidence and reliability of AI model outputs, enabling informed decision-making. This section explores the applications of UQ across various fields, including healthcare, autonomous systems, financial technology, and emerging domains, emphasizing both its transformative impact and the challenges that remain.

\subsection{Healthcare}

In healthcare, the accuracy and reliability of AI-driven predictions are critical due to the potential impact on patient outcomes. UQ serves as a valuable tool to enhance the safety and interpretability of AI systems in two key areas:

\textbf{Medical Imaging.}  AI algorithms have revolutionized medical imaging by automating tasks such as segmentation, classification, and anomaly detection \cite{litjens2017survey}. However, these systems often face challenges due to ambiguous cases, noisy data, or inherent uncertainty in image features. UQ helps address these limitations by providing confidence intervals or uncertainty maps for predictions.

For instance, in tumor segmentation tasks, UQ-enabled models produce pixel-wise uncertainty estimates, highlighting areas where the model predictions are less reliable \cite{jungo2020analyzing}. Such maps allow radiologists to focus on ambiguous regions for further manual analysis. Similarly, in diagnostic support, UQ ensures that predictions with high uncertainty are flagged for clinician review, reducing the risk of diagnostic errors and improving trust in AI systems \cite{faghani2023quantifying}.

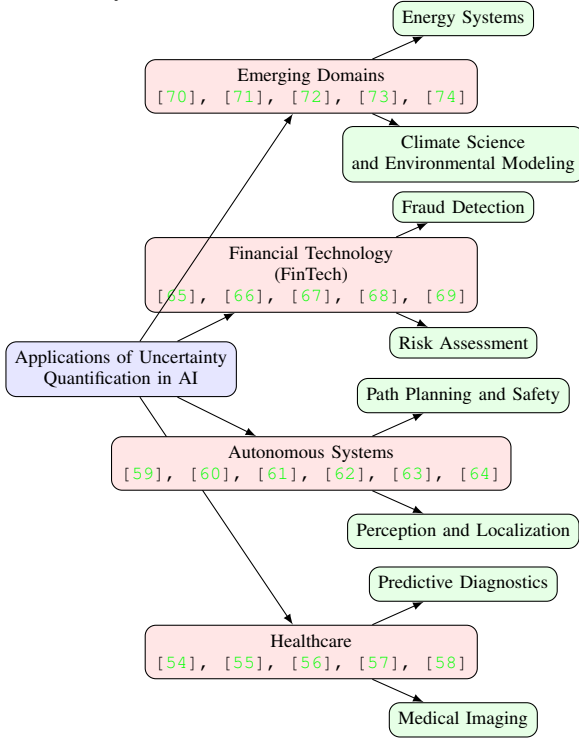
\begin{tikzpicture}[
  grow=east,
  level 1/.style={sibling distance=25mm, level distance=25mm},  
  level 2/.style={sibling distance=18mm, level distance=20mm},  
  edge from parent/.style={draw, -latex},
  every node/.style={font=\scriptsize, align=center}  
]

\node[draw, rectangle, rounded corners, fill=blue!10] {Applications of Uncertainty \\ Quantification in AI}
  child {
    node[draw, rectangle, rounded corners, fill=red!10] {Healthcare \\ \texttt{\cite{litjens2017survey}, \cite{jungo2020analyzing}, \cite{faghani2023quantifying}, \cite{braitman1996predicting}, \cite{dawood2023uncertainty}}}
    child {
      node[draw, rectangle, rounded corners, fill=green!10] {Medical Imaging}
    }
    child {
      node[draw, rectangle, rounded corners, fill=green!10] {Predictive Diagnostics}
    }
  }
  child {
    node[draw, rectangle, rounded corners, fill=red!10] {Autonomous Systems \\ \texttt{\cite{michelmore2020uncertainty}, \cite{wang2023quantification}, \cite{yang2023uncertainties}, \cite{wan2018robust}, \cite{pongpunwattana2004real}, \cite{djuric2020uncertainty}}}
    child {
      node[draw, rectangle, rounded corners, fill=green!10] {Perception and Localization}
    }
    child {
      node[draw, rectangle, rounded corners, fill=green!10] {Path Planning and Safety}
    }
  }
  child {
    node[draw, rectangle, rounded corners, fill=red!10] {Financial Technology \\ (FinTech) \\ \texttt{\cite{behera2023thriving}, \cite{susana2024optimizing}, \cite{marston2018role}, \cite{abdallah2016fraud}, \cite{carson2006uncertainty}}}
    child {
      node[draw, rectangle, rounded corners, fill=green!10] {Risk Assessment}
    }
    child {
      node[draw, rectangle, rounded corners, fill=green!10] {Fraud Detection}
    }
  }
  child {
    node[draw, rectangle, rounded corners, fill=red!10] {Emerging Domains \\ \texttt{\cite{ylhaisi2015quantifying}, \cite{robichaud2020overview}, \cite{quan2019survey}, \cite{srivastava2024distribution}, \cite{lilliu2019uncertainty}}}
    child {
      node[draw, rectangle, rounded corners, fill=green!10] {Climate Science \\ and Environmental Modeling}
    }
    child {
      node[draw, rectangle, rounded corners, fill=green!10] {Energy Systems}
    }
  };

\end{tikzpicture}

\textbf{Predictive Diagnostics.}  Predictive diagnostics leverage AI models to forecast disease risks and patient outcomes. UQ enhances these systems by quantifying the uncertainty in risk predictions, which is crucial when data variability or missing features exist. For example, UQ can stratify patients into risk categories with associated confidence levels, aiding in personalized treatment planning \cite{braitman1996predicting}. In cardiology, uncertainty-aware AI tools can predict the likelihood of cardiac events, enabling proactive intervention while accounting for the variability in patient profiles and sensor measurements \cite{dawood2023uncertainty}.

\subsection{Autonomous Systems}

Autonomous systems, especially in transportation and robotics, operate in complex and uncertain environments. UQ significantly improves the robustness and safety of these systems by quantifying uncertainties in perception, localization, and decision-making processes \cite{michelmore2020uncertainty}.

\textbf{Perception and Localization.}  Autonomous vehicles rely on perception systems to detect and classify objects in their surroundings. These systems are prone to uncertainty due to sensor noise, occlusions, or adverse weather conditions \cite{wang2023quantification}. UQ enables these models to attach uncertainty scores to their outputs, such as bounding box predictions in object detection or segmentation masks \cite{wang2023quantification}. For example, in low-visibility scenarios, UQ can inform the system of reduced confidence in pedestrian detection, prompting caution in vehicle behavior \cite{yang2023uncertainties}. Similarly, localization systems estimate the vehicle's position using sensor fusion techniques. UQ quantifies the confidence in these estimates, ensuring robust navigation in GPS-denied areas or during sensor failures, such as IMU drift or camera obstruction \cite{wan2018robust}.

\textbf{Path Planning and Safety.}  Safe navigation in dynamic environments requires accounting for uncertainties in both the environment and the system's actions. UQ aids in path planning by incorporating uncertainty estimates into the decision-making process \cite{pongpunwattana2004real}. For instance, when predicting the future trajectories of nearby vehicles, UQ helps autonomous systems maintain a safe margin, especially in crowded or unpredictable traffic scenarios \cite{djuric2020uncertainty}. Furthermore, safety-critical tasks like collision avoidance and emergency braking rely on UQ to evaluate the probability of system failure under uncertain conditions, ensuring compliance with stringent safety standards \cite{yang2023uncertainties}.

\subsection{Financial Technology}

In financial technology (FinTech), decision-making often involves high uncertainty due to the dynamic nature of markets and economic systems \cite{behera2023thriving}. UQ has emerged as a critical enabler of robust and interpretable models in this domain.

\textbf{Risk Assessment.}  Risk assessment models in credit scoring and financial forecasting benefit significantly from UQ. Probabilistic models with UQ provide not only point estimates but also confidence intervals, enabling financial institutions to evaluate the reliability of predictions \cite{susana2024optimizing}. For example, in credit scoring, UQ can inform lenders about the uncertainty associated with a borrower's risk profile, helping them make more informed lending decisions \cite{marston2018role}. Similarly, in financial market analysis, UQ quantifies the variability in stock price predictions, allowing investors to optimize their portfolios while accounting for market volatility.

\textbf{Fraud Detection.}  Fraud detection systems often operate in high-stakes environments where false positives and false negatives can have severe consequences \cite{abdallah2016fraud}. UQ-equipped models prioritize transactions for manual review based on their uncertainty scores. For instance, transactions with high uncertainty can signal potentially fraudulent behavior or ambiguous patterns, prompting further investigation while reducing the burden of unnecessary reviews \cite{carson2006uncertainty}.

\subsection{Emerging Domains}

The applications of UQ are expanding into emerging fields, where it addresses unique challenges associated with complex systems and uncertain phenomena.

\textbf{Climate Science and Environmental Modeling.}  
Climate science relies on large-scale models to predict phenomena such as global temperature changes, sea level rise, and extreme weather events. These models are inherently uncertain due to incomplete data, model assumptions, and chaotic system dynamics \cite{ylhaisi2015quantifying}. UQ helps quantify these uncertainties, providing policymakers with confidence intervals for key predictions, such as the likelihood of exceeding certain temperature thresholds under different greenhouse gas scenarios. In environmental modeling, UQ aids in predicting air quality, deforestation rates, and biodiversity loss, ensuring that conservation strategies are based on robust evidence \cite{robichaud2020overview}.

\textbf{Energy Systems.}  In the energy sector, UQ supports the reliable operation of smart grids and renewable energy systems \cite{quan2019survey}. For example, in solar and wind energy forecasting, UQ quantifies the uncertainty in energy generation due to weather variability, enabling grid operators to plan for backup energy sources \cite{srivastava2024distribution}. Similarly, in energy distribution, UQ helps optimize load balancing by accounting for uncertainties in demand and supply predictions, ensuring stable and efficient grid operation \cite{lilliu2019uncertainty}.

\section{Challenges and Future Directions}

\definecolor{lightblue}{RGB}{224, 235, 255}
\definecolor{lightgreen}{RGB}{224, 255, 224}
\definecolor{lightyellow}{RGB}{255, 255, 224}
\definecolor{lightgray}{RGB}{240, 240, 240}

\begin{table*}[htbp]
\centering
\caption{Key Challenges and Future Research Directions in Uncertainty Quantification (UQ)}
\label{tab:uq_challenges_future}
\begin{tabular}{>{\columncolor{lightblue}}p{0.3\textwidth} | p{0.65\textwidth}}
\toprule
\textbf{Key Challenges} & \textbf{Detailed Description} \\
\midrule
\rowcolor{lightgray}
\textbf{Computational Complexity and Scalability} & Many UQ methods, particularly probabilistic techniques like Bayesian inference, are computationally demanding. As AI models become more complex, scalability is essential, especially for real-time applications such as autonomous driving and financial trading. Efficient methods balancing accuracy and computational cost are necessary. \\
\midrule
\rowcolor{lightyellow}
\textbf{Interpretability and Usability} & UQ outputs like confidence intervals and uncertainty decompositions are often difficult for non-experts to interpret, limiting their use in critical fields like healthcare and autonomous systems. Simplifying and visualizing these outputs is key for practical adoption. \\
\midrule
\rowcolor{lightgray}
\textbf{Disentangling and Quantifying Uncertainty Types} & AI systems deal with multiple uncertainties: aleatoric (data noise) and epistemic (model knowledge). In complex tasks like multi-modal learning, these types interact, complicating their disentanglement and impacting decision quality. Properly identifying uncertainty is vital for reliable AI systems. \\
\midrule
\rowcolor{lightyellow}
\textbf{Domain-Specific Constraints} & UQ methods must cater to specific domain constraints. In healthcare, data privacy limits uncertainty modeling, while autonomous systems face real-time and environmental variability. Tailoring methods to each domain's needs is critical for success. \\
\midrule
\rowcolor{lightgray}
\textbf{Ethical and Fairness Concerns} & UQ techniques can amplify biases in model outputs. Poor uncertainty estimates may lead to unfair decisions, such as biased loan approvals. Incorporating fairness into UQ development is necessary to prevent inequitable outcomes. \\
\midrule
\rowcolor{lightyellow}
\textbf{Lack of Standardization and Benchmarks} & The absence of standardized datasets and evaluation metrics hinders method comparison and progress in UQ. Common metrics like calibration error may not be universally applicable, slowing advancements in the field. \\
\bottomrule
\end{tabular}

\vspace{1em}

\begin{tabular}{>{\columncolor{lightgreen}}p{0.3\textwidth} | p{0.65\textwidth}}
\toprule
\textbf{Future Research Directions} & \textbf{Detailed Description} \\
\midrule
\rowcolor{lightgray}
\textbf{Advancing Computational Efficiency} & Developing efficient UQ techniques is essential, with methods like sparse approximations, variational inference, and hardware accelerations (e.g., TPUs). Hybrid approaches combining deterministic and probabilistic methods may strike a balance between accuracy and efficiency. \\
\midrule
\rowcolor{lightgreen}
\textbf{Improving Interpretability} & Enhancing uncertainty visualizations and using explainable AI (XAI) can improve the usability of UQ outputs. Techniques like uncertainty heatmaps or threshold-based alerts can help make uncertainty more understandable and actionable. \\
\midrule
\rowcolor{lightgray}
\textbf{Enhanced Uncertainty Modeling} & Future work should focus on better disentangling aleatoric and epistemic uncertainties, particularly in multi-modal or temporal data environments. Bayesian networks, deep ensemble learning, and causal inference can enhance uncertainty modeling. \\
\midrule
\rowcolor{lightgreen}
\textbf{Domain-Specific Adaptations} & UQ techniques must be tailored to each domain's challenges. In healthcare, integrating clinical expertise can improve diagnostic accuracy. For autonomous systems, real-time uncertainty handling mechanisms are critical for safety. \\
\midrule
\rowcolor{lightgray}
\textbf{Ethical Frameworks for Fair UQ} & Fairness-aware UQ methods, which mitigate biases in uncertainty estimates, should be developed. Ensuring equitable uncertainty estimates across demographic groups is necessary, along with ethical guidelines for responsible UQ deployment. \\
\midrule
\rowcolor{lightgreen}
\textbf{Establishing Benchmarks and Evaluation Standards} & Creating standardized benchmarks with diverse datasets and evaluation metrics is crucial for advancing UQ. These should cover various domains and include synthetic datasets for controlled experimentation, enabling method comparability. \\
\bottomrule
\end{tabular}
\end{table*}

Uncertainty Quantification (UQ) has made significant strides in recent years, establishing itself as an essential component of trustworthy AI systems \cite{abdar2021review}. However, its practical implementation faces numerous challenges that hinder its full adoption across diverse domains \cite{he2023survey,begoli2019need}. These challenges are multifaceted, involving computational limitations, interpretability barriers, the handling of various uncertainty types, and ethical concerns \cite{ovadia2019can,abdar2021review,gawlikowski2023survey,he2023survey}. Overcoming these hurdles requires a concerted effort from the research community, coupled with innovative approaches to drive the field forward \cite{ghahramani2015probabilistic}. This section provides an in-depth discussion of the challenges and outlines future directions for UQ research.

\subsection{Key Challenges}

\textbf{Computational Complexity and Scalability.} Many UQ methods, especially probabilistic approaches such as Bayesian inference and sampling-based techniques, are computationally expensive \cite{neal2012bayesian, gelman1995bayesian}. These methods often involve iterative processes, high-dimensional integrations, or large ensembles of models, all of which demand significant computational resources. As AI models grow in size and complexity, especially in domains like natural language processing and generative modeling, the scalability of UQ techniques becomes a critical concern \cite{blei2003latent, vaswani2017attention}. Real-time applications, such as autonomous driving and financial trading, further exacerbate these challenges, requiring methods that balance accuracy and computational efficiency \cite{kendall2017uncertainties, koza1994genetic}.

\textbf{Interpretability and Usability.} The outputs of UQ methods, such as predictive distributions, confidence intervals, or epistemic-aleatoric uncertainty decompositions, are often difficult for non-expert users to interpret \cite{gal2016dropout, kendall2018multi}. This limits their utility in decision-critical domains like healthcare and autonomous systems \cite{begoli2019need, mcallister2017concrete}. For instance, a radiologist might find it challenging to translate uncertainty estimates from a model into actionable diagnostic insights \cite{ghoshal2020estimating}. Similarly, in autonomous vehicles, presenting actionable uncertainty information to onboard decision systems remains an unresolved issue \cite{feng2020deep, abdar2021review}. Improving the clarity, relevance, and presentation of uncertainty outputs is essential for practical adoption.

\textbf{Disentangling and Quantifying Uncertainty Types.} AI systems encounter multiple forms of uncertainty. Aleatoric uncertainty arises from inherent noise or variability in the data, while epistemic uncertainty stems from a lack of model knowledge or training data \cite{der2009aleatory, gal2016dropout}. In complex tasks, such as multi-modal learning or reinforcement learning in dynamic environments, these uncertainty types often interact, making disentanglement difficult \cite{kendall2018multi, dearden1998bayesian}. Mischaracterizing one type of uncertainty for another can lead to suboptimal decision-making and undermine trust in the AI system \cite{abdar2021review, bhatt2021uncertainty}.

\textbf{Domain-Specific Constraints.} UQ methods must be tailored to the unique requirements and limitations of specific application domains. In healthcare, for example, privacy concerns restrict access to large, high-quality datasets, complicating the development of robust uncertainty models \cite{kaissis2020secure, sheller2020federated}. In autonomous systems, environmental variability and real-time constraints challenge the reliability of uncertainty estimates \cite{kendall2017uncertainties, feng2020deep}. Meanwhile, financial systems must balance uncertainty estimation with strict regulatory and risk management requirements \cite{sim2004robust, hull2012risk}. Addressing such domain-specific constraints is critical for effective implementation \cite{abdar2021review, bhatt2021uncertainty}.

\textbf{Ethical and Fairness Concerns.} UQ techniques are not immune to the biases present in training data or model designs \cite{mehrabi2021survey}. Poorly calibrated uncertainty estimates can perpetuate or amplify biases, leading to inequitable outcomes \cite{o2017weapons}. For example, biased uncertainty estimates in loan approval systems may disproportionately disadvantage certain demographic groups. Ethical considerations, including fairness and transparency, must be integral to the development and deployment of UQ methods \cite{doshi2017towards}.

\textbf{Lack of Standardization and Benchmarks.} The field of UQ lacks standardized datasets and evaluation metrics for consistent benchmarking of methods \cite{gawlikowski2023survey}. While certain metrics like calibration error and sharpness are widely used, they are not always suitable for all tasks or domains \cite{kuleshov2018accurate}. The absence of standardized benchmarks limits comparability between techniques, slowing the pace of progress \cite{hendrycks2019benchmarking}.

\subsection{Future Directions}

To address these challenges, future research in UQ should focus on the following areas:

\textbf{Advancing Computational Efficiency.} Developing computationally efficient UQ techniques is a top priority. Methods such as sparse approximations, variational inference, and low-dimensional projections can significantly reduce the computational burden \cite{hensman2013gaussian, blei2017variational}. Leveraging hardware accelerations, such as tensor processing units (TPUs) and parallel computing, can further improve scalability \cite{jouppi2017datacenter}. Additionally, hybrid approaches that combine the strengths of deterministic and probabilistic methods may offer a balanced trade-off between accuracy and efficiency \cite{wang2016bayesian}.

\textbf{Improving Interpretability.} Incorporating uncertainty visualizations and explainable AI (XAI) techniques can make UQ outputs more accessible to end-users \cite{christoph2020interpretable}. For instance, overlaying uncertainty heatmaps on medical images or using trajectory uncertainty bands in path-planning systems can provide intuitive insights \cite{mcallister2017concrete}. Simplifying complex metrics into actionable thresholds or alerts can further enhance usability in real-world applications \cite{amodei2016concrete}.

\textbf{Enhanced Uncertainty Modeling.} Future research should focus on disentangling aleatoric and epistemic uncertainties more effectively, particularly in complex environments with multi-modal or temporal data \cite{kendall2017uncertainties}. Techniques like Bayesian neural networks, deep ensemble learning, and causal inference can aid in providing a more comprehensive understanding of uncertainty sources \cite{lakshminarayanan2017simple, neuberg2003causality}. Moreover, expanding the scope of UQ to include distributional shifts, adversarial robustness, and uncertainty propagation across model hierarchies is critical for building more robust systems \cite{ovadia2019can, mkadry2017towards}.

\textbf{Domain-Specific Adaptations.} Tailoring UQ techniques to domain-specific needs is essential for adoption. In healthcare, incorporating clinical knowledge into uncertainty estimation frameworks can improve diagnostic accuracy \cite{ghoshal2021estimating}. For autonomous systems, designing real-time uncertainty handling mechanisms for dynamic environments can enhance safety \cite{kahn2017uncertainty}. In financial technology, integrating regulatory requirements with UQ methodologies can ensure both compliance and performance.

\textbf{Ethical Frameworks for Fair UQ.} Developing fairness-aware UQ methods that mitigate biases in uncertainty estimates \cite{kusner2017counterfactual} is a key research direction. For instance, regularizing models to produce equitable uncertainty estimates across demographic groups can address fairness concerns \cite{kallus2022assessing}. Additionally, ethical guidelines and standards for auditing UQ outputs \cite{mittelstadt2016ethics} can help ensure responsible deployment.

\textbf{Establishing Benchmarks and Evaluation Standards.} Creating comprehensive benchmarks that include diverse datasets, uncertainty scenarios, and evaluation metrics is essential for advancing UQ research \cite{ovadia2019can}. These benchmarks should span multiple domains and include synthetic datasets with known uncertainty characteristics for controlled experimentation \cite{gal2016dropout}. Standardizing metrics, such as calibration and sharpness, across tasks will enable more meaningful comparisons between methods \cite{kuleshov2018accurate}.

\section{Conclusion}

Uncertainty Quantification (UQ) is a cornerstone in the advancement of reliable, robust, and interpretable AI systems, underpinning their safe and effective deployment across critical domains. This work has comprehensively reviewed the theoretical foundations, cutting-edge methodologies, and diverse applications of UQ in areas such as healthcare, autonomous systems, financial technology, and emerging fields like climate science and energy systems. Despite notable progress, UQ faces significant challenges, including computational inefficiency, limited interpretability, and difficulties in handling multi-modal and dynamic data. Moreover, the lack of standardized benchmarks and the growing demand to address ethical implications further underscore the need for continued research. Future efforts must focus on developing scalable, interpretable, and domain-adaptive UQ methodologies while integrating them with emerging paradigms like federated learning and quantum computing. Establishing robust evaluation benchmarks and fostering fairness-aware approaches will also be essential for building equitable and trustworthy AI systems. As AI technologies continue to permeate high-stakes environments, mastering uncertainty remains pivotal in shaping systems that are not only intelligent but also dependable, fair, and aligned with societal expectations.

\bibliographystyle{IEEEtran}  
\bibliography{references}

\end{document}